# Linguistic laws in biology

Stuart Semple[1*], Ramon Ferrer-i-Cancho[2*] and Morgan L. Gustison[3*]

[1] School of Life and Health Sciences, University of Roehampton, London, UK
[2] Complexity and Quantitative Linguistics Laboratory, Laboratory for Relational Algorithms, Complexity, and Learning Research Group, Departament de Ciències de la Computació, Universitat Politècnica de Catalunya, 08034 Barcelona, Catalonia, Spain
[3] Department of Integrative Biology, University of Texas at Austin, Austin, TX, USA
* All authors contributed equally

**Corresponding author**: Semple, S. (s.semple@roehampton.ac.uk)

## Abstract

Linguistic laws, the common statistical patterns of human language, have been investigated by quantitative linguists for nearly a century. Recently, biologists from a range of disciplines have started to explore the prevalence of these laws beyond language, finding patterns consistent with linguistic laws across multiple levels of biological organisation - from molecular (genomes, genes and proteins) to organismal (animal behaviour) to ecological (populations and ecosystems). We propose a new conceptual framework for the study of linguistic laws in biology, comprising and integrating distinct levels of analysis - from description to prediction to theory building. Adopting this framework will provide critical new insights into the fundamental rules of organization underpinning natural systems, unifying linguistic laws and core theory in biology.

**Highlights**

• Linguistic laws refer to statistical patterns shared across human languages. Investigation of these patterns has been extended to a range of biological systems, from molecules to organisms to ecosystems, with the number of studies increasing in recent years.

• Linguistic laws and established concepts in different fields of biology may describe similar - or even identical - patterns, providing an opportunity for unification of natural and language sciences.

• We propose an overarching framework which shifts the emphasis from exploring and describing linguistic laws, to identifying underlying mechanisms, generating predictions, and ultimately developing general theory about the organization of biological systems.

• This potential to develop new theory for understanding the natural world will only be realised through cross-fertilisation of ideas between researchers working in diverse disciplines and focussed on different levels of biological organisation.

# The laws of language – from linguistics into biology

Investigations of linguistic laws, the common statistical patterns of human language, have been a cornerstone of quantitative linguistics since the first such laws were proposed in the 1930's [1–3]. More recently, these laws have attracted attention from biologists, who are beginning to explore their prevalence beyond human language (Table 1, Figure 1). Work in this area initially focused on Zipf's rank-frequency law (more often known as Zipf's law), which states that the relative frequency of a word is inversely proportional to its frequency rank [4]. In recent years, there has been a notable increase in research on two further linguistic laws - Zipf's law of abbreviation, which describes the tendency of more frequently used words to be shorter [4], and Menzerath's law, which states that the larger the linguistic construct, the smaller its constituent parts [2,5,6]. A wide range of further linguistic laws exists, and many are yet to be investigated beyond language (e.g. [7]).

Biologists have tested for patterns consistent with linguistic laws among diverse taxa and different levels of biological organisation - molecular [8–17], organismal [18–27] and ecological [28–31] (see Table 1). This work necessitates more general formulations of these laws, which originally were framed in specific linguistic constructs. Additionally, researchers need to reconsider not just the appropriate comparable elements of analysis (e.g. call types, instead of words) but also the relevant dimensions to be measured (e.g. call type duration, instead of word length in letters). The expansion of studies of linguistic laws beyond language raises the question of whether we should rename these phenomena - e.g. to biological laws - to reflect their broader applicability. However, we feel it is appropriate for now to retain the original terminology, in recognition of the field in which the laws were first derived.

Here, we advocate a new approach to exploring linguistic laws in biology, which moves investigations from the level of description to the levels of explanation, prediction and theory building.

## Studies of linguistic laws in biological systems

### Zipf's rank-frequency law

The relationship between the frequency of a word and its rank (1st most frequent, 2nd most frequent etc.) that is expected by Zipf's rank-frequency law [4] (Figure 1B), is traditionally explored in written texts and spoken languages [32,33]. Rank and frequency are by definition negatively associated, and linguists studying this law (or the closely related modification, Zipf-Mandelbrot law – see Table 1) focus on characterizing the degree of linearity and steepness (exponent) of the log-log slope between these measures. In written English, the relationship is highly linear with an exponent ~1 [34], and a study of this law across 100 typologically diverse languages found a narrow exponent range centred around 1 but with some variation (0.76-1.44) [35]. Studies of Zipf's rank-frequency law beyond language come from different levels of biological organisation and diverse taxa (Figure 1C, Table 1).

At the molecular level, frequency distributions consistent with this law (or its sister law, Zipf's number-frequency law – Table 1), are documented in systems including oligonucleotides in DNA sequences [9], secondary structures of RNA [10], gene expression [36] and the size of protein superfamilies [37]. At the organismal level, most work focuses on animal communication, with patterns predicted by this law (or Mandelbrot's modification) documented in the vocalisations of birds (e.g. [21,38–40]) and of both terrestrial and marine mammals [41–43], as well as in primate gestural communication [20]. Beyond the organismal level, Zipf's rank-frequency law describes abundance distributions of both species and communities. For example, patterns consistent with this law are found in the spatial distribution of land snails, *Vallonia pulchella* [28], and the relative abundance of species within a broad range of plant and animal communities [44]. This law has also been applied to epidemiological data, for example to explore patterns of geographic distribution of COVID-19 [30]. In sum, Zipf's rank-frequency law is a notable example of how linguistics laws can apply to biological systems beyond language.

Zipf's law of abbreviation

A significant negative relationship between word length and frequency of use, as expected under Zipf's law of abbreviation [1] (Figure 1B, Table 1), is found across hundreds of typologically diverse languages in written form [45], as well as in spoken [46–48] and sign language [49]. Beyond language, and in marked contrast to Zipf's rank-frequency law, studies that explicitly test Zipf's law of abbreviation have been conducted only at the organismal level - in the behaviour (and particularly vocal communication) of birds and mammals - with the number of studies increasing noticeably in recent years (Figure 1C). A negative relationship with frequency of occurrence has been found for the length of song bouts (measured as the number of calls) in black-capped chickadees, *Poecile atricapillus* [50], call type duration in Formosan macaques, *Macaca cyclopis* [51], vocal phrase size (in component units) in Indri, *Indri indri* [52], gesture duration in chimpanzees, *Pan troglodytes* [23] and the size (in component units) of surface behavioural patterns of dolphins, *Tursiops truncatus* [24]. However, not all species tested follow Zipf's law of abbreviation, at least in the initial analyses [23,53,54]; these examples provide important insights into where and how this law should be explored.

It is important to note that the absence of studies explicitly exploring Zipf's law of abbreviation at molecular and ecological levels does not mean that the patterns predicted by the law have not been investigated at those levels. For example, studies of global size-density relationships in ecology have documented a negative association between population density and body size of species [55] - more frequently occurring species tend to be smaller. This finding highlights a key point: that linguistic laws and established concepts in different fields of biology may describe similar, or even identical, patterns. Identifying such common ground provides a foundation for unification across disciplines, and for the development of more general theory about the organisation of biological systems.

Menzerath's law

Patterns consistent with Menzerath's law - a negative relationship between the size of the whole and the size of the constituent parts [5] (Figure 1B, Table 1) - occur across many languages and linguistic levels (e.g. phonemes, syllables, words, clauses and sentences [56,57]). This law is documented not just in written texts, but also in spoken [47,48] and signed language [58]. However, some studies fail to find the patterns predicted by Menzerath's law [59,60], suggesting it is not as widespread in human language as Zipf's law of abbreviation. Such inconsistencies may reflect variation in how the 'whole' and 'constituent parts' are connected; it is thought that these linguistic levels should be immediately consecutive (e.g. clauses and phrases), rather than distant (e.g. clauses and words) [61]. Investigations of Menzerath's law (or the closely related Menzerath-Altmann law – Table 1) in biological systems initially focused on genes, genomes and proteins, but over the last five years there has been growing interest in applying this law to different forms of animal communication (Figure 1C). At the molecular level, studies demonstrate a negative relationship between exon number and size in genes [14,62], domain number and size in proteins [16], segment number and size in RNA [17], and chromosome number and size in genomes [8,15]. In animal communication, patterns consistent with Menzerath's law are seen in primate vocal sequences [26,52,63–65] and bird song [22,27], as well as in the gestural communication of chimpanzees [23]. However, vocal sequences of a number of primate and bird species do not follow this law [27,66].

Although there is an absence of explicit tests of Menzerath's law at the ecological level, patterns consistent with this law have in fact been documented there. For example, studies of cross-community scaling relationships show a negative association between the total number of individuals in an ecological community and the average size of an individual in that community [55] – the larger the whole, the smaller the constituent parts. This result again highlights the existence of common ground between linguistic laws and - on the surface, completely unrelated - core concepts in other fields.

Other linguistic laws

In addition to work exploring these three linguistic laws in biological systems, a small number of studies investigated other linguistic laws beyond language (Table 1). Patterns consistent with Herdan's law, which describes the relationship between the number of unique words in a text and the overall size of the text, exist in the proteomes of viruses, Archaea, bacteria and Eukarya [13]. Zipf's meaning-frequency law, which states that more frequently occurring words tend to have more meanings, is found in chimpanzee gestures [25] and bottlenose dolphin whistles [19]. Additional tests of these two linguistic laws, and others that have yet to be explicitly tested in biological systems (see [7] for examples), offer opportunities to explore commonalities between human language and diverse natural systems. As described above, it is also possible that patterns predicted by linguistic laws have already been investigated in biological systems but without explicit reference to (and perhaps without knowledge of) one of these laws.

Conditions for the investigation of linguistic laws in biological systems

Linguistic laws are inherently simple in design (Figure 1A and B). Those described above share one core criterion: that a data structure contains at least one group of discrete units. This is the only criterion for Zipf's rank-frequency law and Herdan's law, which explore the distribution patterns of categorically distinct units [4,7,67]. Zipf's law of abbreviation requires, in addition, that units are measured by some aspect of their size, while Zipf's meaning-frequency law requires that units have one or more distinct meanings (or proxies thereof). Menzerath's law involves a hierarchical data structure of two or more unit levels, where larger constructs are made up of units measured by size. All these criteria are broadly applicable across biological systems, which are generally composed of discrete parts and organized in a hierarchical structure.

Beyond the satisfaction of basic criteria, a key issue remains in the exploration of certain linguistic laws in biological systems: what are the most appropriate 'currencies' to use in place of the specific linguistic measures from the original formulation of these laws? In studies of Zipf's law of abbreviation and Menzerath's law in animal communication, for example, duration of call or gesture types is often assessed, but this approach most likely reflects that duration is the easiest - rather than the most appropriate - measure of unit size. A study of rock hyrax, *Procavia capensis*, vocalisations indicates that the choice of currency can be critical [68]. Among males, a negative relationship was found between call type duration and frequency of use, in line with Zipf's law of abbreviation, but among females this relationship was positive (i.e. opposite to the law). However, when call amplitude rather than duration was explored, both sexes conformed to the law – more frequently used call types were quieter. These results highlight that simple, discrete measures of size may not always be appropriate or sufficient. Instead, broader and perhaps multidimensional measures of magnitude, or more precise indices of energetic cost, may be more suitable.

The need to identify a suitable currency may limit the range of systems in which particular linguistic laws can be explored. For example, laws of word meaning can be tested in those animal communication systems where signal types can be assigned a specific meaning (e.g. chimpanzee gestures [25]), and perhaps also in genetic systems where, for example, different combinations of codons 'mean' (i.e. translate into) different amino acids. However, there seems no obvious analogy for meaning at an ecological level, so extending such laws to populations and ecosystems may not be feasible. Ultimately, the appropriate currency to use when testing linguistic laws in biology should be informed by a strong underpinning theory, from which clear predictions are derived to describe the expected patterns and how these patterns are to be assessed.

<u>The universality of linguistic laws, and the role of 'exceptions'</u>

A key issue around the investigation of linguistic laws, in human language and in other systems, relates to the notion of universality. In traditional linguistics [69], two types of 'universals' are typically considered. Absolute universals describe exception-less patterns, while statistical universals describe patterns occurring significantly above chance levels. The latter are the universals that should be considered in relation to linguistic laws, but in studies of these laws beyond human language the distinction between these two types of universals is perhaps not so readily recognised. Finding patterns that do not fit those predicted by linguistic laws - 'exceptions' - is a common theme (e.g. [54,66,70]). However, such exceptions are certainly not unexpected for statistical universals, and can in fact be important for highlighting alternative ways or parts of the system where laws can be tested.

For example, a study of Zipf's law of abbreviation in four bat species found a negative correlation between duration and frequency of production for short range social calls, but not for distress calls [71]. Similarly, while Bezerra et al [54] found no link between duration and frequency of use of call types in the full vocal repertoire of common marmosets, *Callithrix jacchus*, a subsequent analysis by Ferrer-i-Cancho and Hernández-Fernández [53] found conformity to the law in a subset of the repertoire comprising close range social calls. These studies indicate that investigations of linguistic laws should consider patterns that might occur in subsets of the whole repertoire. Developing stronger theory around linguistic laws will allow for specific predictions to be made about where (i.e. in which subsets) we might expect to find conformity to these laws.

A study of Menzerath's law in gorilla, *Gorilla gorilla*, vocal sequences [72] highlights another consideration - whether we should include situations where a constituent part represents the whole construct. In this study, the predicted negative relationship between call duration and sequence length (in terms of the number of constituent calls) was found, but only when analyses included sequences consisting of a single call. When analyses considered only sequences of two or more calls, the negative relationship no longer remained, and in fact a weak positive association was found. Future studies of Menzerath's law should consider data both with and without sequences of length one, to explore this phenomenon further (notably, studies of Menzerath's law in language often exclude words of one syllable [48,60]).

# Studying linguistic laws in biology: a new framework

As investigations of linguistic laws in molecular biology, organismal biology and ecology increase in number (Figure 1C), it is important now to ask what these laws mean in different systems, why these patterns occur, and how we can move the field forward to provide the most important insights. To this end, we propose a new framework for the study of linguistic laws in biology (Figure 2) that builds on the foundations and aims of the scientific method [73,74]; this framework is underpinned by distinct analytical levels and the integration of research across levels. To date, most studies fit into the lower levels of this framework, involving exploratory and descriptive statistics (Level 1) and mathematical modelling (Level 2) or (less often) inferential and predictive modelling (Level 3). There have also been insightful attempts at theory construction for specific laws or study systems, but extension to the creation of more general theory (Level 4) is typically lacking. An important goal for future research will be to shift from the level-centric approach adopted thus far, to bottom-up and top-down integrative approaches where empirical research informs, and is guided by, generalized theory and explicit hypotheses.

Level 1. Exploratory and descriptive statistics: The most common approach to investigate linguistic laws in biological data involves the description and exploration of the patterns these laws predict. At the simplest level, researchers have described qualitatively how well the observed data align with the patterns predicted by a specific linguistic law (e.g. [38]). Stronger, quantitative approaches (Box 1) fit curves or distributions to data, or test for predicted correlations between variables (examples are found in Table 1). Description and exploration are critical for understanding when and where patterns consistent with linguistic laws appear, and identifying exceptions to laws allows these to be explored at higher levels in the framework (see Level 3). However, this approach on its own is unable to answer basic questions about how and why laws emerge in the first place. Such questions are well-suited to mathematical modelling approaches that describe and explain law-like patterns.

Level 2. Descriptive mathematical modelling: This approach is oriented towards describing simple processes that reproduce law-like patterns. A common focus of linguistic law research at this level involves stochasticity. This stems from early observations that Zipf-like word frequency distributions arise from random processes that combine subunits to form units (i.e. Miller's random-monkey model) and generate new text (i.e. Simon's model) [60,75–78], and that other stochastic processes (e.g. random breakage) also reproduce patterns that look like Menzerath's law [79]. Such studies call into question the meaningfulness of linguistic laws, but these arguments are refuted by multiple lines of evidence (see Box 2). A complementary line of work tests whether the expression of law-like patterns by these stochastic models accurately captures patterns in real-word data [60,80–82]. On the whole, descriptive modelling is useful for sorting out how law-like patterns can be reproduced by stochastic or deterministic mechanisms. This approach is limited, however, in that it is unable to answer questions about why such patterns emerge.

<u>Level 3. Inferential and predictive mathematical modelling:</u> Mathematical modelling is not always descriptive, but critically can also serve to interpret, explain, and generate predictions (see Box 2). A noteworthy example comes from work combining mathematical models from information theory and language evolution to explain how arbitrary signals become associated with meaning [83]. Mathematical work also links information theory to linguistic laws. Ferrer-i-Cancho et al [84,85] demonstrated a mathematical relationship between Zipf's law of abbreviation and the information theoretic principle of compression. This is the principle of minimizing the average length of an element in a system, for example by inversely aligning the length of elements with their frequency of occurrence. Gustison et al [26] and Ferrer-i-Cancho et al [86] similarly used mathematical models to connect Menzerath's law and Zipf's rank-frequency law to compression. This work, moving beyond the level of description and simple generative processes, strongly suggests that linguistic law-like patterns in certain systems reflect selection for coding efficiency; from here, predictions under alternative hypotheses can be developed and tested.

For example, in animal communication there is a trade-off between signalling efficiency (linked to compression) and the need to accurately transmit information [84]. This leads to the prediction that conformity with these linguistic laws is expected to be strong in short range communication, but less prevalent in long-range signalling where redundancy helps ensure that information is accurately received [26,84]. This prediction has been supported by studies finding that conformity to laws is not seen in analyses (Level 1) of complete repertoires, but is seen in analyses of close range signals [53,71]. Mathematical approaches that generate predictions in this way are powerful (the stochastic generative processes in Level 2 lack this crucial property – see Box 2) and provide the groundwork for building general theory.

<u>Level 4. General theory:</u> Biological data are rife with patterns, but patterns are of limited value without a solid theoretical framework to explain them [87]. An ideal framework, i.e. a scientific theory, is constructed from a set of interconnected principles that are independent from the observable phenomena [74]. With the growing body of exploratory, descriptive, and inferential research on linguistic laws in biology, alongside the recognition that core concepts in certain fields make strikingly similar predictions to those of linguistic laws, we can now begin to build fully-fledged scientific theory. Common themes emerging from different studies provide a key starting point for this endeavour. For example, studies of animal communication and of gene structure suggest that patterns consistent with linguistic laws in these systems reflect evolutionary trade-offs - between the costs and benefits of signal compression in animal communication [26], and between the costs and benefits of structural change in genes [14]. A general theory about the organization of biological systems, that is underpinned by evolutionary selection and that can accommodate these and other trade-offs, would thus be applicable across molecular and organismal levels of biological organisation.

We have developed one such theory, building from an empirical and mathematical exploration of the vocal sequences of geladas, *Theropithecus gelada* [26]. Here, a negative correlation was found between vocal sequence size and the duration of constituent calls, in line with Menzerath's law (Level 1). Sequence lengths were described mathematically as being consistent with a 'memoryless' process (Level 2), and Menzerath's law was interpreted as a prediction of compression via a generalized cost function (Level 3). These findings were combined with previous work on this law in genes, genomes and proteins, to develop the theory (Level 4) that compression - reflecting a trade-off between efficiency of coding and information transmission success - underpins not just animal (including human) communication but also biological information systems in the broadest sense. Such bottom-up development of general theory then facilitates a top-down approach, driving empirical work and hypothesis development at lower levels. For example, James et al [27] built on this general theory to explore how production mechanisms and learning contribute to the emergence of Menzerath's law in bird song, and proposed the hypothesis that ease of motor production underpins compressional organization in this communication system.

An important challenge now is to develop theory that can incorporate exploration of linguistic laws across all biological levels, and thus can unite evolutionary and ecological studies. A vital first step in developing such theory is to reframe linguistic laws in a more abstract way. This abstraction relies on identifying appropriate currencies that are applicable across a wide range of biological systems. Here again, identifying common ground from studies in diverse systems can provide a valuable starting point. For example, the finding that species abundance distributions follow Zipf's rank-frequency law may reflect the differential 'cost of species' [44], where cost reflects the amount of energy required to produce and maintain the organism (e.g. carnivores are 'costly' due to their high trophic level and thus rare). In human and animal communication, patterns consistent with Zipf's rank-frequency law may similarly reflect a differential cost [88] - here not of species but of the various components of the repertoire (for example, in terms of the energetic costs of production or perception).

We propose reframing linguistic laws in terms of a generalised cost function based on energy, and tentatively propose the start of an overarching general theory: that patterns consistent with linguistic laws reflect pressures that shape the allocation of finite energy in a system. The pressures involved will differ markedly between systems, and identifying and exploring the exact underpinning mechanisms will be an important step. At the molecular and organismal levels, evolutionary selection for optimal allocation of constrained energy may be key; then at the ecological level, competition within and between species for the finite available energy in the environment may be critical. Further development and refinement of a general theory of this kind, based on energy and its allocation, is a promising approach to unify all levels of biological organisation. Recent work in molecular biology [89], organismal biology and ecology [90] has similarly focused on energy as a unifying currency, and an exciting prospect is to reconcile general theory around linguistic laws with, for example, metabolic theory of ecology [91] and the equal fitness paradigm [92].

Level integration: Contemporary studies of linguistic laws in biology are largely level-centric, with an emphasis on the descriptive approaches outlined in Levels 1 and 2 above. These lower-level studies are important for developing basic knowledge, but their narrow focus means that we risk being left with a patchwork of datasets and models for different laws and biological systems. To advance understanding, we need more research that connects and integrates levels, achieving a cyclical process (akin to that of the scientific method [74]) in which bottom-up and top-down approaches work in tandem to build, test, and refine general theory (Figure 2). This process should involve research on systems that conform to, or are an exception to (e.g. [54,66,70]), one or more linguistic laws. Exceptions can be used to make clear predictions for testing in independent systems, and in turn, drive theory building and refinement. Collectively, integrative research efforts will promote general theory that joins distinct, unifying principles to understand how, when, and why linguistic laws manifest in biological systems. Such theory will be particularly powerful, allowing the development of clear hypotheses and predictions that are sufficiently abstract in their framing (for example in terms of the system, or units of analysis) as to be applicable across all levels of biological organisation.

## Concluding remarks

Studies of linguistic laws in biology have provided important insights, but the true power of such work has yet to be fully recognised. Many questions remain to be addressed (see Outstanding Questions), and there are opportunities to expand this line of inquiry into new disciplines and towards linguistics laws that have yet to be investigated beyond human language. There is potential to analyse existing biological databases to test linguistic laws, and to explore common ground between such laws and core principles and concepts in different scientific fields. This opens up exciting opportunities for broad scale comparative analyses – not just across taxa but also across diverse biological systems and different levels of biological organisation. Above all, there is an opportunity to reshape how we investigate linguistic laws in biology, integrating different levels of analysis with the ultimate goal of generating and testing general theory. Such unifying work, built on the open exchange and cross-fertilisation of ideas from multiple disciplines, will provide new understanding of the fundamental rules of organization underpinning diverse biological systems – from molecular to organismal to ecological.

**Acknowledgments**

We sincerely thank Daniel Perkins, Logan James, Daniel Takahashi, Antoni Hernandez-Fernandez, two anonymous reviewers and Andrea Stephens for their insightful comments. MLG is supported by the grant K99MH126164 from NIH (National Institutes of Health). RFC is supported by the grant TIN2017-89244-R from MINECO (Ministerio de Economía, Industria y Competitividad) and by the recognition 2017SGR-856 (MACDA) from AGAUR (Generalitat de Catalunya).

## Box 1. Key methodological issues in the study of linguistic laws

Two main statistical approaches are used to investigate linguistic laws: model fitting and correlation. Model fitting, which is the traditional approach taken in quantitative linguistics, involves finding a mathematical function (i.e. an equation) that defines the dependency between two variables. Putative mathematical functions for a number of linguistic laws are shown in Table 1. A key advantage of the model fitting approach is that it helps characterize the non-linear behaviour of a law. Any resulting equations are then tailored by modifying parameters to fit new datasets (e.g. Zipf's rank-frequency exponent varies across languages [35]). On the other hand, these fitted equations are only first approximations to the true, highly complex, relationship between variables [34,93,94]. The validity of many of these equations has been questioned using both theoretical and empirical arguments [32,95–97]. As a reaction to these challenges, as well as a need to generalize laws to diverse systems [84], correlational approaches have become increasingly popular (e.g. [45]).

A correlational approach involves the use of statistical tools to test for positive or negative associations between variables. This approach has some key advantages. First, it simplifies the analysis with respect to curve fitting, and the use of appropriate statistical tools (e.g. Spearman's correlation, linear mixed effect models) removes some assumptions about how variables are structured and related to one another, or controls for the effect of multiple factors. Second, it is supported by theoretical arguments involving optimal coding that predict a negative (or null) correlation consistent with Zipf's law of abbreviation or Menzerath's law [85]. A limitation of correlational analysis is that it is not appropriate for all laws, specifically those where variable definitions are intrinsically related (e.g. frequency can only decrease as rank increases in Zipf's rank-frequency law; type number can only increase with token number in Herdan's law; Table 1). For other laws, it is important to confirm that there are not 'inevitable' correlations due to variables being functionally dependent [79,98]. Researchers have used several strategies to successfully address this criticism for Menzerath's law and Zipf's law of abbreviation [23,26,99].

There are many analytical tools available to explore linguistic laws while avoiding statistical pitfalls. In addition to the references above, we also point readers to discussions on the advantage of maximum likelihood over least squares [100], the use of randomization approaches to develop appropriate null models [27], the problem of rank as a random variable [97], and the problem of independence violation [93].

## Box 2. Debates on the meaningfulness of linguistic laws

Since Zipf's foundational research [4], many researchers have cast doubts on the meaningfulness and utility of linguistic laws. A recurrent criticism is that linguistic laws are inevitable. For Zipf's rank-frequency law, this argument is illustrated with the metaphor of typing, whereby choosing simple units (e.g. letters, nucleotides) at random can result in a data structure conforming to the law [75,78,101,102]. For Menzerath's law, this argument is derived from the definition of the size of the constituent parts as an average [79]: if the total size of a whole construct is constrained, then randomly breaking it into a few versus many pieces will inevitably result in larger versus smaller parts. This line of reasoning also extends to Zipf's law of abbreviation [99]. However, the inevitability of such patterns has been falsified in several ways: (i) by finding patterns across different species and systems that are not consistent with laws (e.g. [23,43,103]); (ii) through mathematical analysis to show that random typing does not in fact reproduce Zipfian laws of unit frequency [104] and that defining constituent part size as an average does not inevitably lead to Menzerath's law [98]; (iii) by showing statistical differences in how laws are expressed in real data as opposed to the artificially constructed datasets used to support the inevitability argument [80,81,105].

Another criticism, applied mainly to Zipf's rank-frequency law but that can be generalized to other laws, is that the presence of a law does not allow inferences to be made about how it works or what function it serves in a specific system [60,106,107]. The main reason for this criticism is that Zipf's rank-frequency law (and others) can be reproduced in many ways, which questions whether a law is meaningful for any specific system. This criticism is an important one, and reflects the current emphasis on exploratory and descriptive studies (Levels 1 and 2 in Figure 2). To thoroughly address this criticism, a shift is needed towards work that makes explanations for laws, develops predictions that can be tested in independent contexts, and builds a general theoretical framework to guide how future studies are designed (Levels 3 and 4). An example of a novel prediction is provided by a model originally designed to reproduce Zipf's rank-frequency law based on cognitively realistic principles; this model sheds light on how children learn new words, while random typing or Simon's model fail to make any prediction [82]. The multiplicity of ways of reproducing a law (Level 2), e.g. [87], does not imply a multiplicity of explanations (Levels 3 and 4).

**Figure legends**

**Figure 1. Common linguistic laws and publishing trends.** (A) An artificial dataset is illustrated using different symbol-colour combinations. There is a repertoire of eight discrete unit types that differ in their relative sizes (left), and the full dataset involves a collection of 64 units that are grouped into aggregates ranging in size from one to ten units (right). For example, in human language the units could be words and the aggregates phrases, while in animal communication the units could be call types in the vocal repertoire and the aggregates vocal sequences. (B) This artificial dataset conforms to patterns expected under Zipf's rank-frequency law (left), Zipf's law of abbreviation (centre), and Menzerath's law (right). The symbols in the Zipfian law plots represent specific unit types, while symbols in the Menzerath's law plot represent aggregates. (C) Cumulative number of papers that explicitly test these three linguistic laws in biological systems, published over the last ten years, assessed from Web of Knowledge using search terms 'Zipf*' or 'Menzerath*' on February 22, 2021. For Zipf's rank-frequency law, also included are the closely related Zipf-Mandelbrot law and Zipf's number-frequency law; for Menzerath's law, also included is Menzerath-Altmann law. Papers are categorized based on the biological level of organization of the study system (molecular, organismal, ecological). Striking differences are clear between the laws in terms of the total number of studies conducted, and in particular the levels of organisation at which each law has been investigated.

**Figure 2. Conceptual framework for investigating linguistic laws in biological systems.**
Research on linguistic laws typically falls into one of four analytical levels. 'Exploratory and Descriptive Statistics' (Level 1) is the most basic level, involving studies that test for conformity to linguistic laws in real world biological systems (the signs of correlations and equations in Table 1). 'Descriptive Mathematical Modelling' (Level 2) involves work that describes processes that reproduce these laws. 'Inferential and Predictive Modelling' (Level 3) involves research that uses computational approaches to explain possible functions of these laws and develops testable predictions. 'General Theory' (Level 4) involves work to build scientific theory by integrating a set of interconnected principles that are independent from the observable phenomena. This framework encourages a research approach that is not centred on a single level, but instead integrates levels through bottom-up (dark blue arrows) and top-down (light blue arrows) approaches.

**Table 1. Key linguistic laws, including all that have been explicitly investigated beyond human language**. For each law, we show the pair of variables related (Var. 1 and Var. 2), mathematical models that have been proposed as a first approximation, and the sign of the correlation between the variables when the law holds. A 'type' is an abstraction for a distinct unit (e.g. a word or a behaviour), $r$ is the frequency rank of a type (the most frequent has rank 1, the 2nd most frequent has rank 2, etc.), $f$ is the frequency of a type, $n$ is the number of types (that have a certain frequency), $l$ is the size of a type (e.g. its length or duration), $\mu$ is the number of meanings or a proxy thereof (e.g. behavioural contexts of a type), $t$ is the number of tokens, $S_w$ is the size of a whole construct (usually its number of parts) and $S_P$ is the size of its parts. * is used to indicate correlations that are inevitable given the definition of the variables involved. In the models, $c$ indicates a positive proportionality factor that, depending on the law, is not a free parameter because it can be deduced before fitting (e.g. applying normalization). $\alpha,\ \beta,\ \gamma,\ \delta$ and $\eta$ are positive constants, the so-called exponents of the laws. $a$ and $b$ are additional parameters.

(For a more comprehensive bibliography on linguistic laws in biology, see: https://cqllab.upc.edu/biblio/laws/)

| Definitions of laws | | | | | Level of biological organisation at which the law has been explored, with examples of studies | | |
|---|---|---|---|---|---|---|---|
| Law | Var. 1 | Var. 2 | Model | Correlation | **Molecular** | **Organismal** | **Ecological** |
| **Zipf's rank-frequency law**. This law is popularly known as Zipf's law but is only one of many laws in Zipf's popular book [4]. $\alpha$ is a positive parameter that is the so-called exponent of the law. $\alpha \approx 1$ for English words. By definition of rank, a negative correlation between $f$ and $r$ is expected for $\alpha{>}0$. | $r$ | $f$ | $f{=}cr^{-\alpha}$ | ≤0* | Codons [108] DNA [9] | Vocal communication [42] Gestural communication [20] | Intra-population abundance distribution [28] Species abundance distributions [29] Disease spread [30] |
| **Zipf-Mandelbrot law**. This is a refinement of Zipf's rank-frequency law by Mandelbrot [109] that consists of introducing an additional parameter $b$. When $b{=}0$, one obtains Zipf's rank-frequency law. | $r$ | $f$ | $f{=}c(r{+}b)^{-\alpha}$ | ≤0* | RNA [10] | Vocal communication [21] | Intra-population abundance distribution [28] Species abundance distributions [31] |
| **Zipf's number-frequency law**. This law and Zipf's rank-frequency law are two sides of the same coin and the relationship between their respective exponents obeys approximately $\beta{=}1/\alpha +1$ [110]. Contrary to Zipf's rank-frequency law, a correlation between $n$ and $f$ is not expected a priori. $\beta$ is a positive parameter that is the so-called exponent of this law. $\beta \approx 1$ for English words [96]. | $f$ | $n$ | $n{=}cf^{\beta}$ | ≥0 | Proteins [37] | | |
| **Zipf's law of abbreviation**. Zipf found that more frequent words tend to be shorter [4] but he did not propose a function for the relationship between $f$ and $l$. In a popular article, Sigurd et al [111] adopted an equation shown here that suggests that word frequency is determined by word | $f$ | $l$ | $f{=}cl^{b}b^{l}$ | ≤0 | | Vocal communication [22] Gestural communication [23] Non-vocal behaviour [24] | |
| length while information theory suggests that it is rather the other way around [85]. See also [112]. | | | | | | | |
| **Zipf's law of meaning distribution.** $\gamma$ is the exponent of this law, that is $\gamma \approx 0.5$ for English words [113,114]. | $r$ | $\mu$ | $\mu=cr^\gamma$ | ≤0 | | Vocal communication [19] | |
| **Zipf's meaning-frequency law.** This law is a prediction by Zipf from the rank-frequency law and the law of meaning distribution [113]. He predicted $\delta$=0.5 for English words, and later it was proven that $\delta = \gamma/\alpha$ [115]. | $f$ | $\mu$ | $\mu=cf^\delta$ | ≥0 | | Vocal communication [19]<br>Gestural communication [25] | |
| **Herdan's law (also known as Heaps' law).** The law defines the growth of the number of distinct words as a function of the text length measured in tokens [67,116]. | $t$ | $n$ | $n=ct^\gamma$ | ≥0* | Proteomes [13] | | |
| **Menzerath's law.** The law is usually defined as a negative correlation between $S_w$ and $S_P$. It bears the name of P. Menzerath [2,5], who was inspired by the relationship between the number of syllables of a word ($S_w$) and the duration of its syllables ($S_P$). | $S_w$ | $S_P$ | - | ≤0 | | Vocal communication [26]<br>Gestural communication [23] | |
| **Menzerath-Altmann law.** A generalization and mathematical formulation of Menzerath's law by G. Altmann [6]. In addition to the proportionality parameter $c$, it is defined by two additional parameters, $a$ and $b$. $a$ is an exponent that usually takes negative values. Although Solé [79] claimed that $a$=-1 and $b = 0$ are inevitable when $S_P$ is defined as a mean size, the argument turned out to be flawed [98]. | $S_w$ | $S_P$ | $S_P = cS_w^a e^{bS_w}$ | ≤0 | Genes [14]<br>Genomes [15]<br>Proteins [16]<br>RNA [17] | | |

## Outstanding Questions

- **Which linguistic laws hold in which biological systems?**

It is vital to explore whether an absence of evidence for linguistics laws in different systems reflects reality or just a lack of exploration; critically, many linguistic laws have never been explicitly tested beyond language, and investigating these could provide novel insights.

- **What underpins conformity to linguistic laws in different biological systems?**

Broad scale comparative approaches will allow testing of hypotheses about the functionality and potential ecological and/or evolutionary drivers of patterns consistent with linguistic laws, and investigation of whether patterns reflect the sharing of a common ancestral trait (constraint) or convergent evolution (selection).

- **What are the most appropriate currencies to use when exploring linguistic laws in biological systems?**

Comparative approaches necessitate the development of measures for testing laws that are applicable across taxa in one system, but also across different systems and levels of biological organization (for example, direct or indirect measures of energetic cost).

- **How does the manifestation of linguistic laws shift across time in biological systems?**

Temporal variation in conformity to linguistic laws is poorly understood; it is important now to explore how these patterns are shaped by both internal processes (e.g., physiological states, ontogenesis) and external conditions (e.g., environmental contexts).

- **Should we rename linguistic laws, and if so how?**

As linguistic laws hold in diverse biological systems, arguably they should be renamed to reflect this broader applicability; any change in terminology should consider the commonality between laws and key concepts in different fields.

- **What are key areas and aims for future studies of linguistic laws in biology?**

Investigations of linguistic laws at levels of biological organisation (e.g. cells, tissues, organs) and in disciplinary fields (e.g. biosemiotics, developmental biology) where they have rarely – or never – been explored (Figure 1) could open up new opportunities for cross-disciplinary integration, and further contribute to development of general theory (Figure 2).

16,

Figure 1

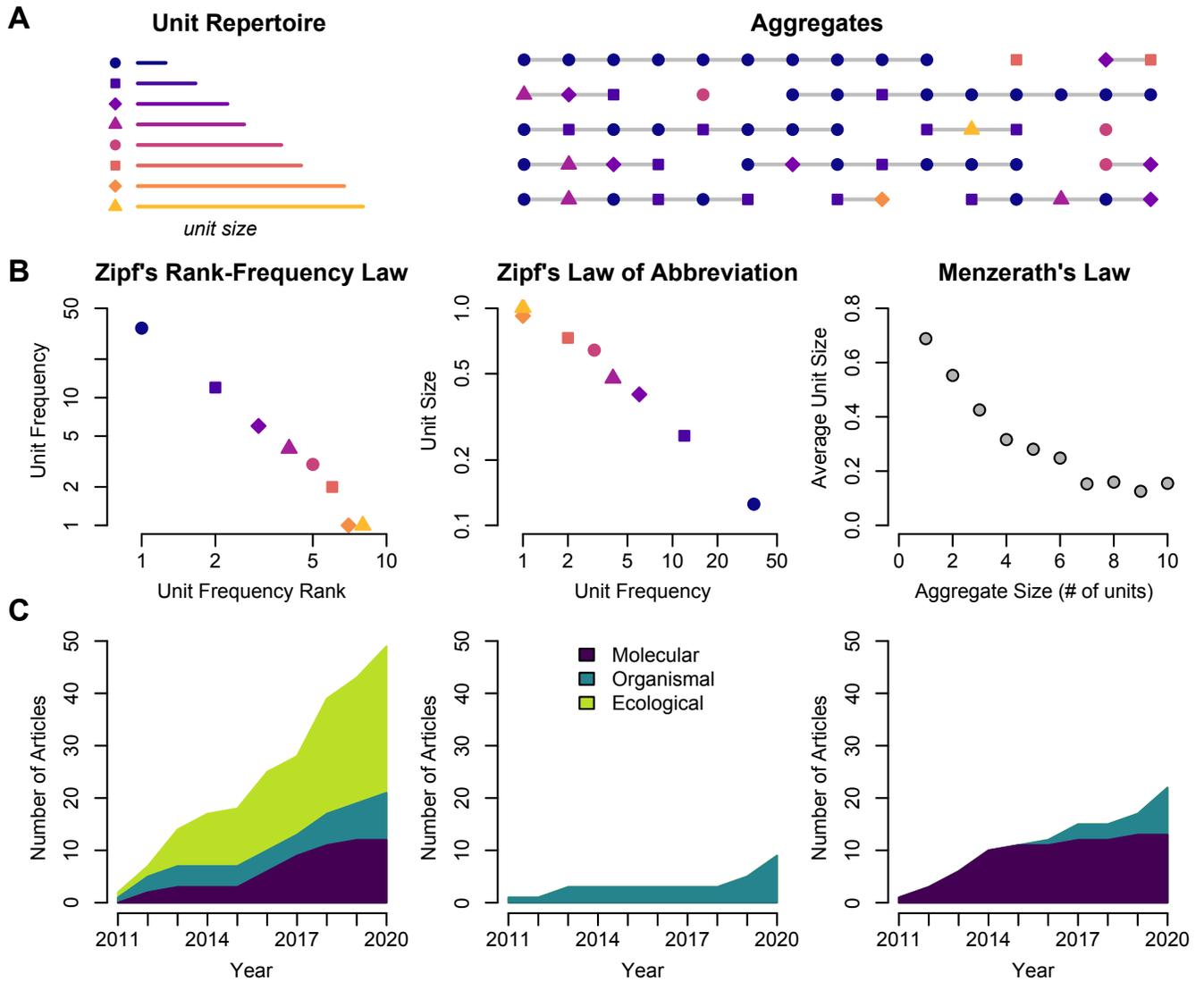

**Figure 2**

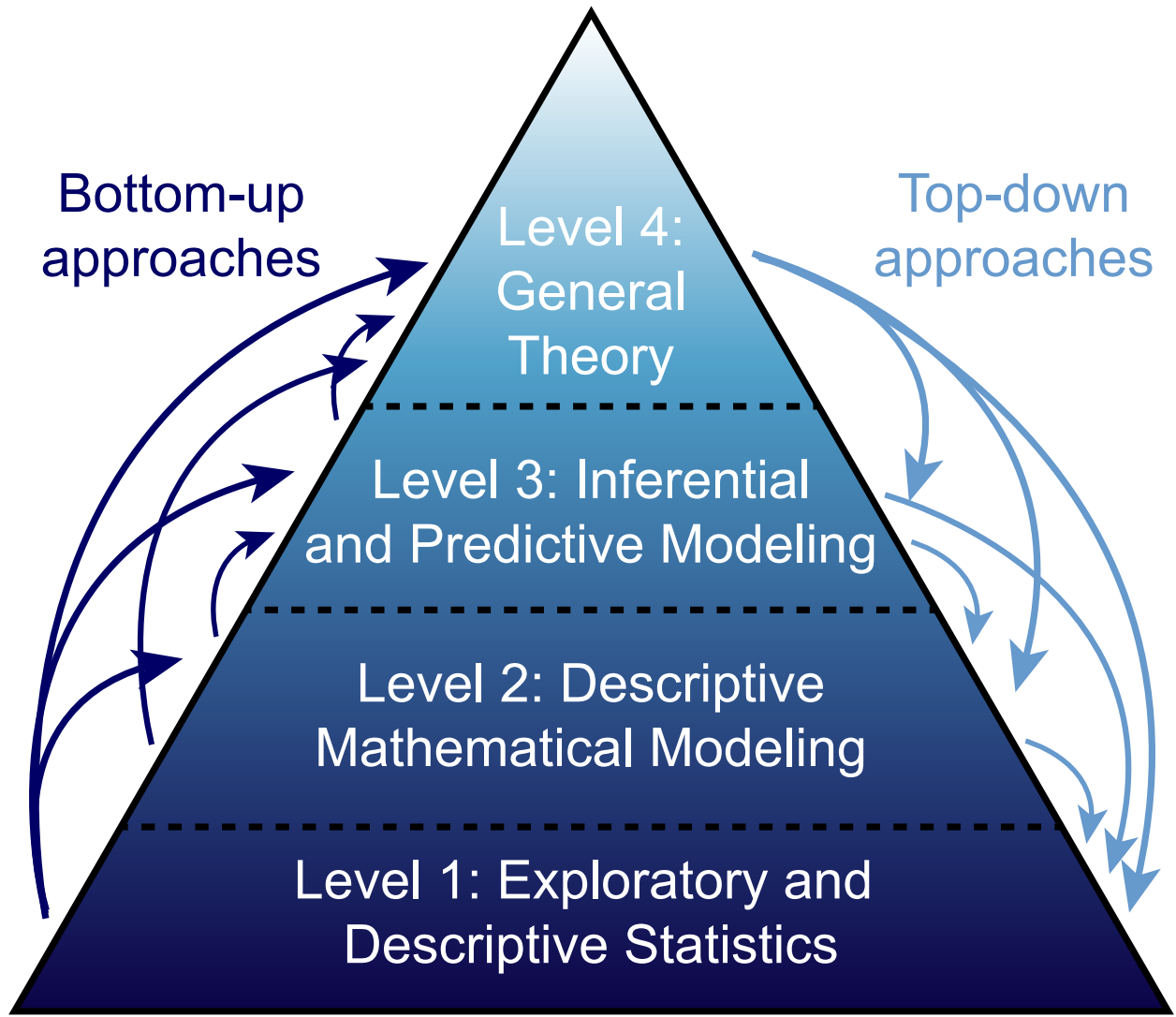